# XGBoost energy consumption prediction based on multi-system data HVAC

Yunlong Li[1],
Wuyi University,
Guangdong，China,
1316827294@qq.com,

Yiming Peng[2],
Wuyi University,
Guangdong, China,
m@163.com,

Dengzheng Zhang [3],
Wuyi University,
Guangdong, China,
m@163.com,

Yingan Mai[4],
Wuyi University,
Guangdong, China,
10@qq.com,

Zhengrong Ruan[5],
Wuyi University,
Guangdong, China,
10@qq.com,

*Shufen Liang,
Wuyi University,
Guangdong, China,
838666003@qq.com

**Abstract**—The energy consumption of the HVAC system accounts for a significant portion of the energy consumption of the public building system, and using an efficient energy consumption prediction model can assist it in carrying out effective energy-saving transformation. Unlike the traditional energy consumption prediction model, this paper extracts features from large data sets using XGBoost, trains them separately to obtain multiple models, then fuses them with LightGBM's independent prediction results using MAE, infers energy consumption related variables, and successfully applies this model to the self-developed Internet of Things platform.

## I. INTRODUCTION

Human beings' pursuits have steadily moved away from sufficient food and clothing and toward a higher quality of life as society has developed. Air conditioning systems with a high energy efficiency ratio are becoming increasingly common as a means of ensuring human comfort. As a result, entrepreneurs place a premium on building thermal comfort. To maintain a comfortable temperature and humidity level, heating, ventilation, and air conditioning systems are needed, despite their high energy consumption. China's building energy consumption accounts for approximately 30% of total society energy consumption, with air-conditioning energy consumption accounting for approximately 60% of total building energy consumption, indicating that air-conditioning energy consumption accounts for nearly 20% of total society energy consumption. Additionally, air conditioning is one of the most energy-intensive components in base stations and communication spaces, accounting for about 45 percent of the communication industry's overall energy consumption. At the moment, central air conditioners in public buildings in China suffer from a variety of issues, including random selection, backward dispatching mode, low automation level, and low operation and maintenance level, all of which are critical for energy conservation and consumption reduction. To save energy and reduce air conditioner consumption, the first step is to forecast its energy consumption, identify the primary energy sources, and facilitate energy-saving transformation. According to the available literature, only a few people use machine learning algorithms to forecast HVAC energy consumption. Louis et al. [1] in the United States cluster various energy consumption parameters and calculation time series using a shape-based approach and the k-medoids algorithm and dynamic time warping to determine similarity. Both Artificial Neural Network (ANN) and Support Vector Regression (SVR) models are tested as representative regression algorithms for a variety of predictor sets. The mean square error is used to evaluate results. Tomasz Szul et al. used the Takagi-Sugeno fuzzy model [4] to forecast the utility of thermally improved building energy efficiency. Gianluca et al. used a building energy model (BEP+BEM) to determine the sensitivity of building environmental parameters, and the results indicated that internal building parameters had a major effect on the building energy consumption page. The second portion of this paper discusses the central air conditioning system and the relationship between the air conditioning components and energy consumption. Additionally, it briefly presents the self-developed Internet of Things framework for energy-saving control, as well as the networking mode and control concept. The third section describes the database creation process and illustrates it with some sample data. These data are obtained in real time by the Internet of Things network. Additionally, attempt to explore and evaluate the data, including the Pearson correlation coefficient between the data for each attribute and the system's power consumption. Among attributes with a correlation coefficient greater than 0.5, and attributes with a correlation coefficient less than or equal to 0.3, a second-order correlation analysis is performed. In the fourth section, we predict XGBoost and

LightGBM separately and then combine them with MAE, and the optimal parameter model is solved using cross-validation, a confusion matrix, and a greedy algorithm.

## II. EXPERIMENTAL SYSTEM AND PLATFORM

This experiment used an Internet of Things platform to monitor and control the central air conditioning system of a university teaching building in South China. This air-conditioning system is nearly identical to the conventional design, consisting of three components: a cooling tower, a computer room, and an air handling unit. The difference is that instead of a horse air conditioner, the air handling device is comprised of a three-phase motor and a volute fan. (see fig.1)

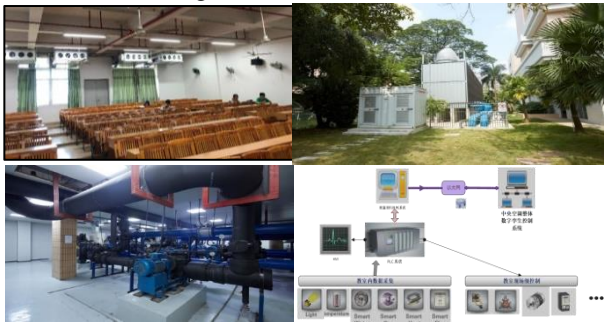

Figure 1 Composition of air conditioning system in a university in South China

A controller is mounted at the terminal to enable automatic cooling control, and the PaaS platform is controlled in the computer room to enable energy-saving control from the refrigeration demand to the refrigeration supply end. Figure 2 depicts the overall system deployment.

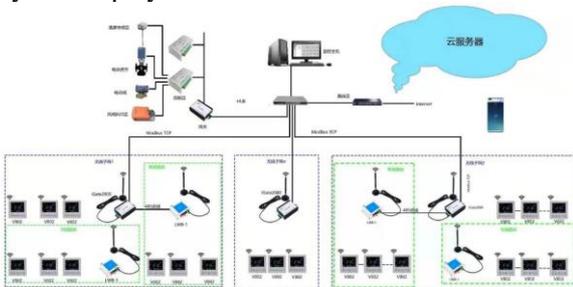

Figure 2 Platform deployment

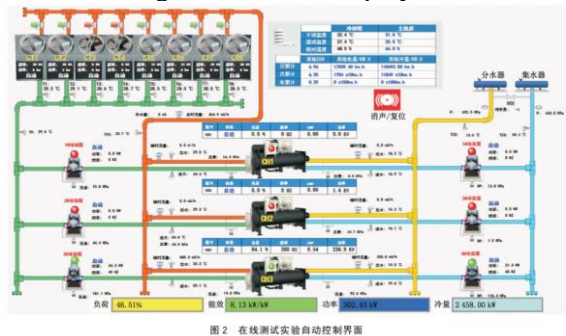

Fig. 3 Interface diagram of refrigeration system

This platform possesses many significant characteristics:

1. It is simple to install: it can be mounted directly on newly installed central air conditioners.
2. Excellent energy-saving effect: energy-efficient central air-conditioning system regulation.
3. The terminal can determine the temperature field in real time: at the terminal, the temperature field can be established using existing real-time sensor data, and the temperature matrix data can be used to change the temperature field for energy savings.
4. The terminal controller should be plug-and-play: in order to deploy the scene, cold source devices such as fans and fans must be controlled, which enables the scene to be easily connected to the platform.

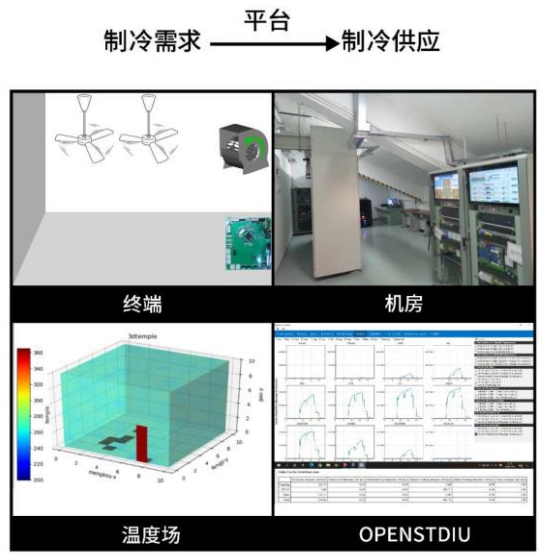

Fig. 4 Effect diagram of platform use

It is important to create a common carrier for the platform of central air conditioning systems in order to simplify product deployment. The gateway connects the central air conditioning equipment's controller to the platform, which sends control instructions to the equipment. At the terminal, the terminal controller developed in this work is used to link and monitor the refrigeration equipment in the scene through the platform.

1. The gateway section: It is primarily used as a connection to connect with the platform's central air-conditioning equipment, and it sends data to the platform following gateway fault detection.

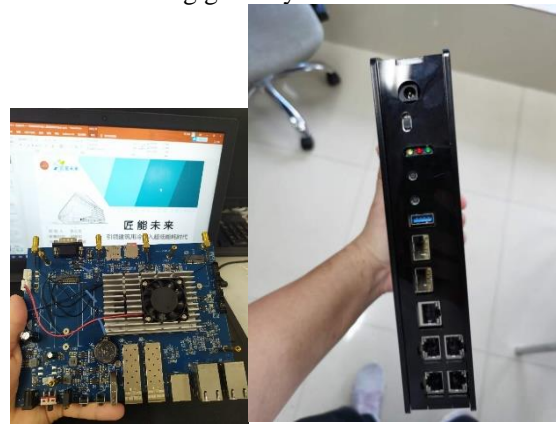

图 1 网关实物图

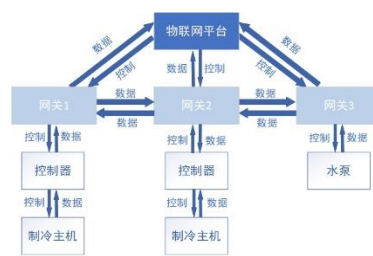

图 2 物联网平台连接示意图

(2) Control component:

(1) control component for the computer room: Given the complexity of central air-conditioning system engineering, it is difficult to optimize the system's operation impact solely by optimizing its local components. It is important to view all components and links of the central air-conditioning system holistically and to organize and monitor the entire system comprehensively. The primary objective of this team's research is to develop an energy-efficient and resource-conserving model of central air conditioning systems combined with various equipment components. The system's parameters are calibrated under normal summer cooling conditions to ensure that it is still in the best possible condition, which offers suggestions for efficiently resolving the "two low and one high" issue with central air conditioning systems.

(1) The central air-conditioning equipment shall create an energy consumption monitoring platform, with the equipment's energy consumption index serving as the evaluation index. 2 The equipment room's dimensions are increased, nonlinear training is conducted, and a control model is obtained. ③ By using Adaboost to pick the required control model, the system's control accuracy and energy-saving efficiency can be improved.

(2) Terminal control part: Since the refrigeration demand of the terminal fluctuates, the terminal controller performs feedback control based on the temperature field, and cold source equipment such as fans and fans is calibrated to bring the temperature field closer to human comfort.

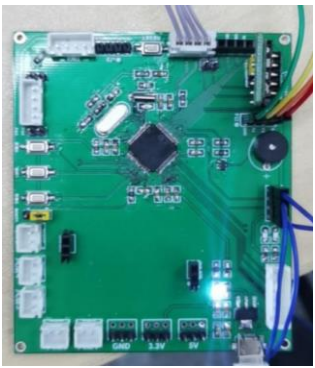

Fig. 7 Terminal temperature controller

(3) Access component:

The Internet of Things platform can dock various types of data, including image data (as shown below), sensor time series data (temperature, operating parameters, $CO_2$ concentration, and so on), and use these data to analyze the cooling capacity of each region and adjust the cooling capacity of the refrigeration device to suit the region's human comfort.

Table 1 Data Transmission by Gateway on March 17, 2021

| 时间 | 1号主机总有功功率 | 机房日累计电量/度 | 机房年累计电量/度 | 机房年平均COP/W |
|---|---|---|---|---|
| 14:12:01 | 11.7 | 224.6 | 365849.2 | 7.1 |
| 14:13:01 | 54.2 | 225.7 | 365849.2 | 7.1 |
| 14:14:01 | 60.8 | 227 | 365849.2 | 7.1 |
| 14:15:01 | 44.8 | 228.4 | 365849.2 | 7.1 |
| 14:16:01 | 44.1 | 229.3 | 365849.2 | 7.1 |
| 14:17:01 | 45.7 | 230.6 | 365849.2 | 7.1 |
| 14:18:01 | 45 | 231.6 | 365849.2 | 7.1 |

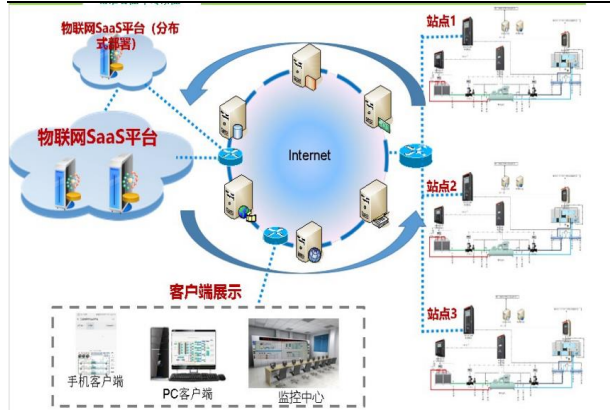

Fig. 9 Schematic diagram of communication

I. CORRELATION ANALYSIS AND FEATURE SELECTION

To collect data, the Internet of Things platform continuously tracks and tests each module of the central air conditioning system. The database is MySql, and the protocol is mqtt. It includes real-time data on the main engine, cooling tower, cooling system, refrigeration pump, and electric valve, as well as power usage during service and standby days. Due to the fact that the air conditioning system operates 365 days a year, the refrigeration system is not always operational, and therefore is more likely to operate during the summer. As a result, we take the lead in preprocessing the database, using the system's regular cumulative cooling ability as the object, filtering out non-zero data, i.e., the refrigeration system's usual operating data, and extracting one month's (March) data as the training data set for independently training XGBoost and LightGBM models. While training the model, collect data from the following month's air conditioning systems (May) as a test data set for verification of the results. To improve the data set's training, it's necessary to first analyze the raw data. Pearson correlation coefficient analysis is performed on the preprocessed data to determine the relationship between the data for each attribute and the system's power consumption. Among attributes with a correlation coefficient greater than 0.5, and attributes with a correlation coefficient less than or equal to 0.3, a second-order correlation analysis is performed. We extract nine function vectors: (1) Temperature of the cooling water inlet to the No. 2 main engine (2) Regular accumulated power in the machine room (3) No. 1 main engine cooling water outlet temperature (4) No. 1 main engine cooled water outlet temperature (5) No. 1 main engine cooling water inlet temperature (6) Temperature of the outdoor wet bulb (7) Machine room power consumption in real time (8) Cooling water outlet temperature of No.3 main engine (9) Cooling water outlet temperature 1.

the attributes of some data can be extracted from the attributes of another data or a group of data.

$$r = \frac{\sum_{i=1}^{n}(x_i - \bar{x})(y_i - \bar{y})}{\sqrt{\sum_{i=1}^{n}(x_i - \bar{x})^2}\,(y_i - \bar{y})^2}$$

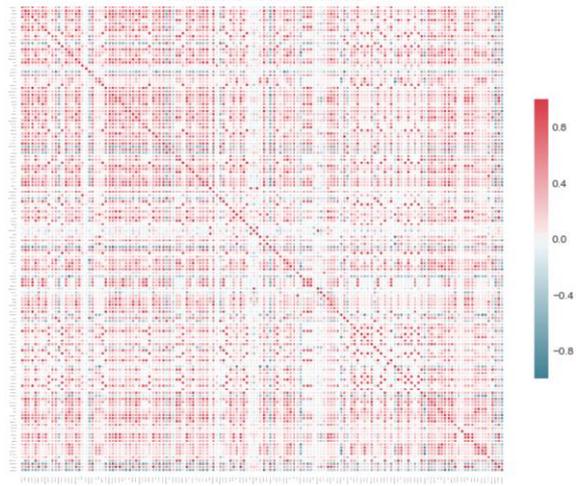

Fig. 12 correlation coefficient display

The closer the absolute value is to 1, the stronger the correlation between the operating parameters, and the closer it is to zero, the weaker the correlation. According to research [6], when |r| exceeds 0.5, the two parameters are substantially correlated, while when |r| is less than 0.3, they are just marginally correlated.

## II. XGBOOST MODEL AND LIGHTGBM MODEL ARE MERGED ACCORDING TO MAE

A. independent prediction via XGBoost

Mr. Chen Tianqi of Washington University created XGBoost[2] in 2016 as an extensible machine learning framework. XGBoost is not a model strictly speaking, but a software package that enables users to easily solve classification, regression, and sorting problems. It internally implements the Gradient Lifting Tree (GBDT) model and optimizes the algorithm within the model, resulting in high accuracy at an extremely fast speed. The heart of XGBoost is its integration concept, which entails parallel processing, or the operation of multiple models. The realization method entails constructing an objective function, optimizing it approximately, incorporating tree structure into the objective function, and optimizing it using a greedy algorithm. XGBoost is implemented on an internet of things platform, and the operation database of a cooling tower with variable flow and complete operating condition serves as the data set for this model, which includes the main engine, cooling tower, cooling system, refrigeration pump, electric valve, operation power consumption, and standby

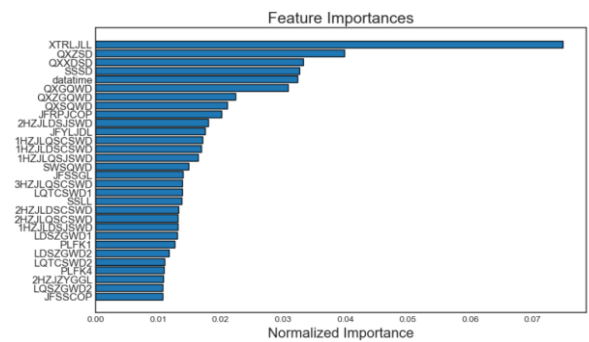

Fig. 10 feature selection

Table 2 The database accepts some data display every day

| | 主机日总有功功率/W | 冷冻泵总有功功率/W | 冷却塔总有功功率/W | 机房日累计电量/° | 系统日累计冷量 |
|---|---|---|---|---|---|
| 2021-03-16 | 40.5 | 4 | 0 | 225.8 | 2315.6 |
| 2021-03-17 | 49.2 | 4 | 0 | 223.7 | 2304.7 |
| 2021-03-18 | 52.6 | 4 | 0 | 227.8 | 2323.4 |
| 2021-03-19 | 0 | 3.9 | 0 | 224.1 | 2313.8 |
| 2021-03-20 | 44.2 | 3.9 | 0 | 223.9 | 2306.4 |
| 2021-03-21 | 0 | 0 | 0 | 223.6 | 2303.5 |
| 2021-03-22 | 0 | 0 | 0 | 224.0 | 2308.9 |

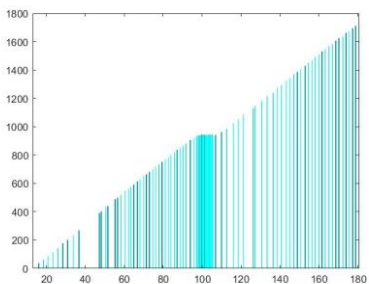

Fig. 11 Correlation between daily power consumption and daily cooling capacity

We use Pearson correlation coefficient [5] to represent the correlation between any two variables when there are a large number of collected data and power consumption. In this article, XGBoost is used to forecast the energy usage of an air conditioning system. The results of multi-model interaction are directly summed and compared to the true value using residual training, and the minimum residual is obtained by continuous optimization.

III. the process of constructing an objective function

The overall energy consumption of the air conditioning system is measured and forecast under the assumption that the air conditioning design conditions and platform regulation meet the energy conservation requirements. The database includes information about the main engine, cooling tower, cooling pump, refrigeration pump, electric valve, running and standby power usage, which are all registered in real time each day. Due to the magnitude, this article abstracts these forms of energy consumption as k trees. k=7 in this experiment. The objective function is as follows as a result of the combined operation of k trees:

$$Obj = \sum_{i=1}^{n} l(y_i, \hat{y}_i) + \sum_{k=1}^{K} \omega f(k)$$

In which:

$$\hat{y}_i = \sum_{k=1}^{k} f_k(x_i), f_k \in \beth$$

$\sum_{i=1}^{n} l(y_i, \hat{y}_i)$ is referred to as the loss function, and it refers to the loss incurred by n samples.

And $\sum_{k=1}^{K} \omega f(k)$ is referred to as a standard concept, and its aim is to regulate complexity. The standard term is composed of the number of leaf nodes t and the square of the leaf score's L2 modulus.

$$\omega(ft) = \gamma T + \frac{1}{2}\mu \sum_{j=1}^{T} \omega_j^2$$

XGBoost optimizes the objective function and pre-prunes it.

(2) Superposition training

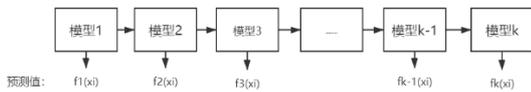

Given $x_i$, there are

$$\hat{y}_i^{(0)} = 0$$
$$\hat{y}_i^{(1)} = f_1(x_i) = \hat{y}_i^{(0)} + f_1(x_i)$$
$$\hat{y}_i^{(2)} = f_1(x_i) + f_2(x_i) = \hat{y}_i^{(1)} + f_2(x_i)$$
$$......$$
$$\hat{y}_i^{(n)} = f_1(x_i) + f_2(x_i) + ... + f_n(x_i)$$
$$= \hat{y}_i^{(n-1)} + f_n(x_i)$$

(3) Approximating the objective function with Taylor series Finally, assuming that the tree's form is understood, the following functions are optimized:

$$\sum_{j=1}^{T}\left[\left(\sum_{i \in I_j} g_i\right) \cdot W_j + \frac{1}{2}\left(\sum_{i \in I_j} h_i + \mu\right) \cdot W_j^2 + \alpha T\right]$$

(4) Find the parameters

$$W_j^* = -\frac{G_j}{H_j + \mu}$$

$$Obj^* = -\frac{1}{2}\sum_{j=1}^{T}\frac{G_j^2}{H_j + \mu} + \alpha T$$

(5) Adjusting parameters

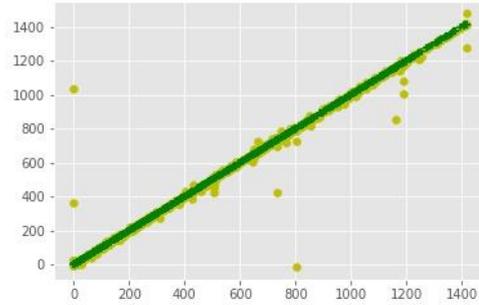

Fig. 13 Fitting diagram of XGBoost single prediction results

B. forecast independently using LightBGM [1]

LigthGBM is a new member of the boosting set model family that is as efficient at realizing GBDT as XGBoost. In general, it is similar to GBDT and XGBoost in that it fits a new decision tree using the negative gradient of the loss function as the residual approximation of the current decision tree.

LightGBM will outperform XGBoost in a variety of ways. It is more efficient in terms of training time and precision, consumes less memory, supports parallel learning, large-scale data processing, and makes direct use of category features. LightGBM makes use of the histogram algorithm, which consumes less memory and simplifies data separation. The concept is to discretize continuous floating-point features into k discrete values and then create a Histogram with a width of k. Then iterate through the training results, accumulating the cumulative statistics for each discrete value in the histogram. When selecting functions, we simply need to traverse in order to find the best segmentation points based on the histogram's discrete values. The computing cost is then significantly reduced. The pre-ranking algorithm computes the split benefit for each eigenvalue traversed, while the histogram algorithm computes it k times (k can be considered constant), and the time complexity is reduced from

O(#data*#feature) to O(k*#features). LightGBM employs a leaf-wise growth strategy, in which it selects the leaf with the highest splitting benefit (generally the most data) from all current leaves and splits it, and so on. As a result, when compared to Level-wise, Leaf-wise will eliminate more errors and achieve higher accuracy while maintaining the same split times. The downside of Leaf-wise is that it can result in the formation of a large decision tree and over-fitting. As a result, LightGBM applies a maximum depth cap to Leaf-wise in order to ensure high performance and avoid over-fitting.

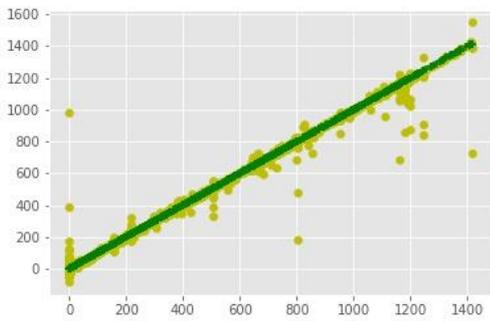

Fig. 14 Fitting diagram of prediction results using LightGBM alone

C. Integration according to MAE

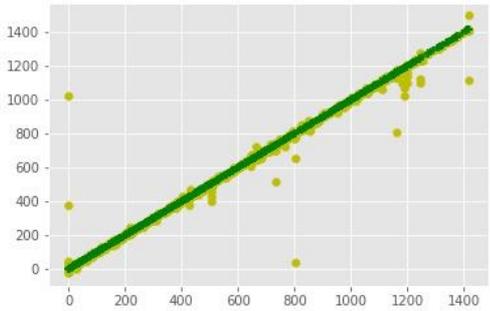

The weight of XGBoost is 0.7100118044045896, and the weight of LightGBM is 0.28998819559541045

### IV. VERIFICATION AND COMPARISON OF RESULTS

A. comparison between A. XGBoost and LightGBM

|  | XGBoost | LightGBM | 融合模型 |
|---|---|---|---|
| Accuracy | 0.840533 | 0.845123 | 0.850137 |
| Memory consumption | 4.853GB | 0.822GB | 1.786GB |
| Time cost comparison | 4520 | 478 | 475 |

B. Cross-validation

Cross-fundamental validation's concept is to divide the original data into two parts; one is used as a training set, while the other is used as a verification set. To determine the classifier's accuracy, the classifier is trained using the training set and then checked using the verification set.

### V. SUMMARY

The air-conditioning system of a university in south China is used as an experimental system in this paper, and the energy usage is estimated using a fusion of the XGBoost and LightGBM models on a self-developed Internet of Things platform. The results obtained using this approach have a high degree of fitting and are more accurate than those obtained using a support vector machine SVM. Accurate energy consumption prediction results will assist engineers in carrying out effective energy-saving transformations, improving device testing, and ensuring the system's continued health.

ACKNOWLEDGEMENT

The author Yunlong Li thanks Yiming Peng for help in experiments, Guanghua Zhou for idea contribution and discussions and Shufeng Liang for fruitful discussions. In addition, we also thanks to the free resources provided by Wuyi university.

REFERENCES

[1] Guolin Ke, Qi Meng,Thomas Finley. LightGBM: A Highly Efficient Gradient Boosting Decision Tree [M]2016.
[2] Tianqi Chen. XGBoost: A Scalable Tree Boosting System [M]2021.
[3] Dhowmya Bhatt, Danalakshmi D. Forecasting of Energy Demands for Smart Home Applications [M]2020
[4] Tomasz Szul, Krzysztof N˛ecka. Application of the Takagi-Sugeno Fuzzy Modeling to Forecast Energy Effifiiciency in Real Buildings Undergoing Thermal Improvement Takagi-Sugeno [M]2017
[5] 杨辉.基于运行参数选择的中央空调系统主要设备能耗估计方法研究[D].重庆大学,2016.
[6] Buda A, Jarynowski A. Life time of correlations and its applications. [Vol. 1][M]. ABRASCO - Associação Brasileira de Saúde Coletiva, 2010.